# Matching Underwater Sonar Images by the Learned Descriptor Based on Style Transfer Method

Xiaoteng Zhou
School of Ocean Engineering
Harbin Institute of Technology
Weihai, China
zhouxiaoteng@stu.hit.edu.cn

Changli Yu*
School of Ocean Engineering
Harbin Institute of Technology
Weihai, China
* Corresponding author: yuchangli@hitwh.edu.cn

Xin Yuan
School of Ocean Engineering
Harbin Institute of Technology
Weihai, China
xin.yuan@upm.es

Citong Luo
School of Ocean Engineering
Harbin Institute of Technology
Weihai, China
luocitong@gmail.com

*Abstract*—This paper proposes a method that combines the style transfer technique and the learned descriptor to enhance the matching performances of underwater sonar images. In the field of underwater vision, sonar is currently the most effective long-distance detection sensor, it has excellent performances in map building and target search tasks. However, the traditional image matching algorithms are all developed based on optical images. In order to solve this contradiction, the style transfer method is used to convert the sonar images into optical styles, and at the same time, the learned descriptor with excellent expressiveness for sonar images matching is introduced. Experiments show that this method significantly enhances the matching quality of sonar images. In addition, it also provides new ideas for the preprocessing of underwater sonar images by using the style transfer approach.

## I. INTRODUCTION

The development of marine resources has gradually become one of the hot topics today. In the process of autonomous underwater vehicles (AUVs) exploration in the deep sea, due to the absorption and scattering of light by seawater, optical sensing technology is almost ineffective. In contrast, acoustic sensing technology can resist turbidity and weak light in seawater [1], so sonar equipment based on acoustic sensing technology has become the only sensor that can detect long-distance scenes and finish target classification and target recognition on the seabed. In tasks such as simultaneous localization and mapping (SLAM), side scan sonar (SSS) performs well [2,3]. In some researches on the underwater sonar image matching, many scholars also conduct research based on SSS images [4-6]. On the other hand, due to the limitations of sonar equipment, the imaging process will encounter problems such as low signal-to-noise ratio (SNR), low resolution, low feature repeatability, and the imaging edges are blurred, there are serious distortions, and the overall image quality is poor. In the sonar images, the visual content of the same target captured from various directions is also very different [4,7]. These deficiencies in imaging will affect the performances of the sonar image matching task. Therefore, how to effectively preprocess the underwater sonar images is still a thorny and long-term problem.

Taking into account that traditional feature descriptors like scale invariant feature transform (SIFT) [8,9] are developed for optical images, and there are problems of extraction difficulties, low efficiency and inaccuracy on the sonar image matching task so that they perform generally [4,10]. In order to better solve this limitation, we adopt two methods in this article: The first is to make the sonar image style as close to the optical one as possible. The second is to introduce some more expressive feature descriptors based on the convolutional neural networks (CNN). Based on the first idea, for these non-traditional optical field image matching tasks, some researchers try to use the style transfer method to solve this problem [11,12]. The classic style transfer idea proposed by Gatys [13] has made great achievements in the field of computer vision. Its purpose is to use the advantages of CNN's deep feature extraction to distinguish the content structure and texture style of the image, and then combine images of various contents and styles to achieve style transfer purpose. Among them, Jang [12] used the style transfer method to effectively achieve the matching between underwater acousto-optic images. The innovation in this paper is also inspired by them.

However, in the actual ocean exploration project, due to the perspective of the sensor imaging differences and respective limitations, it is hardly to obtain sonar and optical images with very similar visual attributes for the same target. Secondly, once an image with an optical style has rich content, the content of the optical image will inevitably be learned in the process of style transfer, which will cause conflicts with the original content of the sonar image and affect the matching task. Therefore, this article starts from the practical engineering requirements and considers that when the SSS is used to perform seabed search operations, the targets faced are rare, and most of which are useless monotonic backgrounds. We focus on one of the main target images of sonar. In the matching study between image pair, the style transfer method based on factor control is selected [14], which can change the image style according to the three factors of spatial positioning, color and illumination information, and cross-space scale.

In brief, we can control the style transfer process in a targeted manner according to the characteristics of the current sonar image. The purpose is to find a balance between maximizing the original features of the sonar images and optimally matching between sonar image pairs. Compared with the method cited in [12], the flexibility of the algorithm is fully improved, and it also helps us to obtain high-quality and clear underwater sonar images with optical style.

Recent studies have proved that the learned descriptors based on CNN framework [15-17], after training on a specific optical dataset, the matching characteristics are significantly better than the traditional local feature descriptors, among which the TFeat [17] is a local feature descriptor based on CNN, with low computational overhead and low network structure complexity. On the premise of ensuring matching accuracy, it has significant advantages in matching speed, so it is very suitable for applications in large-scale underwater detect tasks.

To preserve the content of the sonar image and make it having an optical style so as to better play the role of TFeat descriptor, which is trained on optical datasets, we preprocess sonar images by controlling the process factors of transfer. This paper explores the performances of the learned descriptors based on the CNN framework in underwater sonar image matching tasks. We focus on the real-time and accuracy requirements of AUV large-scale operations.

The main contributions of this letter are as follows:

- Preprocessing sonar image using style transfer method
- Evaluating the quality of the generated image
- Using the TFeat descriptor for sonar image matching

The pipeline of our proposed method is depicted in Fig. 1. First input two sonar images and then introduce an optical image based on the visual characteristics of the current sonar images, and then they simultaneously transfer the optical style and perform control adjustments in feature extraction stage. Next, match features on the generated images, and finally map the feature points with the original sonar images. (The SSS image on the left side of the Fig. 1 is obtained by Deep Vision AB company [18] using the DeepEye 680D in Lake Vättern, Sweden.)

## II. METHOD

### A. Transfer framework

The core of the image style transfer method [13] is to use the CNN network to find the optimal matching problem of content and style. The key point of the transfer is how to separate the content representation and style representation in a CNN network.

The high-layer features of the network are generally information such as the object and layout of the image, and the low-layer features generally express the pixel information of the image. In other words, when extracting image content features, the visual expression effects of various layers are different. The features extracted at the lower layer will retain more detailed information of the original image, however the upper layer will mainly retain the semantic information. The loss in content is:

$$L_{content}\left(\vec{p},\vec{x},l\right) = \frac{1}{2}\sum_{i,j}\left(F_{ij}^l - P_{ij}^l\right)^2 \quad (1)$$

where $\vec{p}$ denotes the input sonar image, $\vec{x}$ denotes the generated image, $F_{ij}^l$ is activation of the $i^{-th}$ filter at position $j$ in layer $l$.

The style features $G^l$ are obtained by solving the Gram matrix on the basis of $F^l$. The matrix is mainly used to explain the texture features of the image. The specific calculation is:

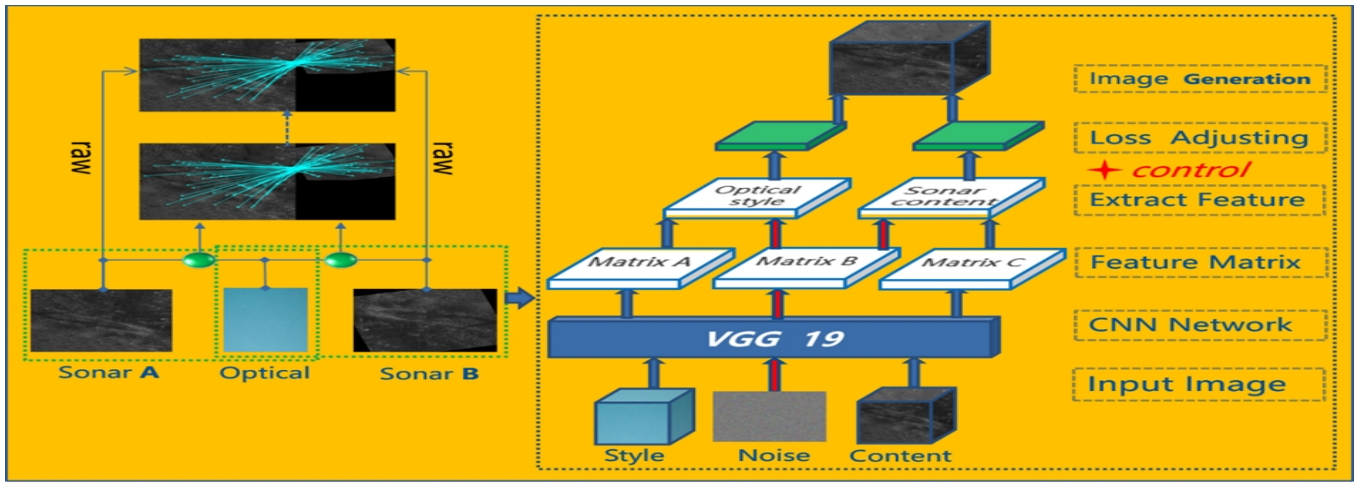

Figure 1. Pipeline of our proposed method.

$$G_{ij}^{l} = \sum_{k} F_{ik}^{l} F_{jk}^{l} \quad (2)$$

Among them, $i$ and $j$ represent the $i^{-th}$ feature map and $j^{-th}$ feature map of the layer, and $k$ represents the $k^{-th}$ element of the feature map, so formula (2) is to calculate the inner product of the two feature maps and it is independent of their positions.

$$E_{l} = \frac{1}{4N_{l}^{2}M_{l}^{2}} \sum_{i,j} \left( G_{ij}^{l} - A_{ij}^{l} \right)^{2} \quad (3)$$

$$L_{style}\left(\vec{a}, \vec{x}\right) = \sum_{l=0}^{L} w_{l} E_{l} \quad (4)$$

where $\vec{a}$ denotes original image and $\vec{x}$ denotes generated image, $A^l$ and $G^l$ their respective style representation in layer $l$, $w$ is the weighting factor, used to explain the importance of the features of each layer. The $G^l$ of each layer of $x$ will calculate the mean square error (MSE) together with the $A^l$ of each layer of $a$. It is calculated according to the weighting factor $w$. This loss is used to describe the difference of style. The final loss function of the style transfer method is calculated as follows:

$$L_{total} = \alpha L_{content} + \beta L_{style} \quad (5)$$

where $\alpha$ and $\beta$ are the custom parameters, which represent the respective weights of content reconstruction and style reconstruction.

### B. Factor control mode

In the process of style transfer [14], three assumptions were made for factor control: (I) different areas have various styles (II) colors are related to content (III) the content structure is divided into fine and rough parts, and they are independent. It is mainly divided into color and illumination information control, spatial positioning control and cross-space scale control. From the perspective of overcoming the limitations of sonar images and the quality of transfer, we focus on the analysis of color and illumination information control.

*1) Luminance-only transfer:* The control effect of lighting information is mainly reflected in the brightness. By adopting the method of brightness preservation, the image is converted from RGB mode to YIQ mode, and the IQ with the color information is kept unchanged, then only the brightness conversion is performed on the Y channel. Considering the low brightness of the sonar image itself, if we try to transfer a high-brightness style image, it needs to match the mean and variance of its brightness to the sonar image. The specific working principle is as follows:

$$L_{s'} = \frac{\sigma_{C}}{\sigma_{B}} \left( L_{S} - \mu_{S} \right) + \mu_{C} \quad (6)$$

where $\mu_S$ and $\mu_C$ are the mean luminances of the two images, $\sigma_S$ and $\sigma_C$ are their standard deviations.

*2) Color histogram matching:* Firstly, we utilize the color conversion algorithm to match the style image with the color of the content image, and define it as a new image $N$, and then use $N$ and the image providing the content as the input of the neural style to transfer. The working principle is as follows:

$$P'_{S} = Ap_{S} + b \quad (7)$$

this is a linear transformation, where $A$ is a 3×3 matrix, $b$ is a 3-vector. Use this to make adjustments so that mean and covariance of the RGB values in new style image $P'_S$ match those of target image.

### C. Learned descriptor TFeat

The TFeat is based on CNN's local feature descriptor which is obtained from a new approach that utilizes triplets of training samples in the learning process. Its loss design breaks the traditional idea of exploiting pairs of positive and negative patches to learn discriminative CNN representations, which greatly reduces the complexity of the network structure, and has big advantages in speed performance.

*1) Learning with triplets:* It involves training from samples of the form $\{a, p, n\}$. $a$ and $p$ are different viewpoints of the same physical point, and $n$ comes from a different keypoint. By optimizing the network parameters, $a$ and $p$ are close in the feature space, while $a$ and $n$ are separated. Introduce two parameters $\delta_+$ and $\delta_-$, the formula for both is as follows:

$$\begin{cases} \delta_{+} = \left\| f(a) - f(p) \right\|_{2} \\ \delta_{-} = \left\| f(a) - f(n) \right\|_{2} \end{cases} \quad (8)$$

By combining triplet with the following two losses, we explore their performances in patch pair classification and nearest neighbour matching application.

*a) Margin ranking loss:* This is a convex approximation to the [0-1] ranking error loss, which mainly used to evaluate the violation of the ranking order of embedded features within the triplet, where $\mu$ is a margin parameter.

$$\lambda(\delta_{+}, \delta_{-}) = \max(0, \mu + \delta_{+} - \delta_{-}) \quad (9)$$

*b) Ratio loss:* In [19], it studied the ratio loss, which optimizes the ratio distance within the triplet. Ratio loss learns follow embeddings like $\frac{\delta_{-}}{\delta_{+}} \to \infty$, where the ranking loss forces the embeddings to be learned.

$$\hat{\lambda}(\delta_{+}, \delta_{-}) = \left( \frac{e^{\delta_{+}}}{e^{\delta_{+}} + e^{\delta_{-}}} \right)^{2} + \left( 1 - \frac{e^{\delta_{-}}}{e^{\delta_{+}} + e^{\delta_{-}}} \right)^{2} \quad (10)$$

If there is no margin related to the loss, for all values of $\delta_+$ and $\delta_-$, we have $0 \le \hat{\lambda} \le 1$.

The procedure of the underwater sonar images matching enhancement algorithm can be expressed as follows.

---
**Algorithm1**: Underwater sonar images matching enhancement algorithm (**USME**)

**Input**: Custom optical image *C* and two sonar images *A* and *B*
**Output**: Matching result of *A* and *B*
**Preprocessing**:
  Feed *A* and *C* to VGG19 and get content of *A* and style of *C*
  Feed *B* and *C* to VGG19 and get content of *B* and style of *C*
**For** *i* <= *num_Iteration* **do**
  | Generate image *A'* with content of *A* and style of *C*
  | Generate image *B'* with content of *B* and style of *C*
**End**
**Matching**:
  Match *A'* and *B'* with the learned descriptor TFeat
  Mapping correspondences to *A* and *B*
**End**

---

## III. EXPERIMENTAL DATA SETTING

We select six sonar images for testing, which were taken by PENG CHENG LABORATORY using SSS based on real underwater scenes [20]. We divide them into three image pairs.

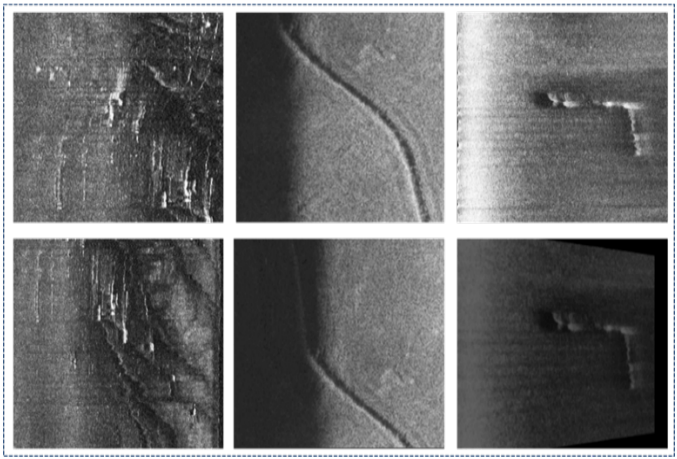

Figure 2.  Underwater sonar images.

TABLE I.  CHARACTERISTICS OF DETECTED OBJECTS

| Sonar image | Image detail | |
|---|---|---|
| | *Object* | *Imaging feature* |
| Image pair 1 | Sea rock | general side scan imaging |
| Image pair 2 | Pipeline | shadow obstruction |
| Image pair 3 | Swale | brightness and viewing angle difference |

We select two optical images (leftmost column, top to bottom 1,2) to provide the optical style for the sonar objects (from the left to the right at the top of the image is a, b and c). Among them,1 is from datasets used in the reflection and texture studies from Columbia University [19], and 2 is the natural glass appearance. See Fig. 3 for details.

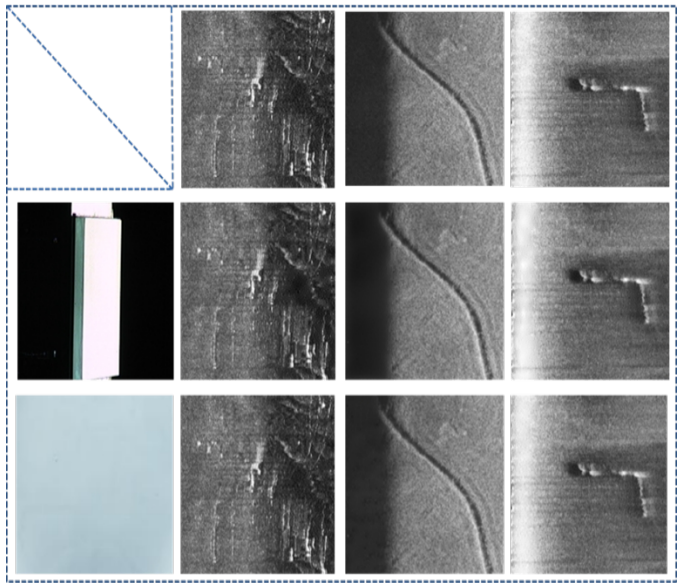

Figure 3.  Raw sonar images and sonar images generated after style transfer.

In Fig. 3, we randomly select an image from each of the three sets of image pairs as the content, then use styles above to transfer, the top row is the raw picture. Finally, take peak signal-to-noise ratio (PSNR), structural similarity index matrix (SSIM), cosine similarity (COSIN), information entropy (Infro-EN) four indexes [21] to evaluate the quality of generated images. They are all used to evaluate two styles of transfer, the optimal one will get 10 points. The style with the highest total score is used for subsequent matching task.

TABLE II.  GENERATED IMAGES QUALITY EVALUATION

| Image | The evaluation index | | | | Total score |
|---|---|---|---|---|---|
| | *PSNR* | *SSIM* | *COSIN* | *Infro-EN* | |
| Style1[a] | 39.2581 | 0.9928 | 0.9886 | **7.1929** | 10 |
| Style2[a] | **48.3461** | **0.9986** | **0.9995** | 7.1475 | 30 |
| Style1[b] | 32.4056 | 0.9622 | 0.9611 | **7.0228** | 10 |
| Style2[b] | **46.5058** | **0.9876** | **0.9883** | 6.8857 | 30 |
| Style1[c] | 40.1019 | 0.9877 | 0.9771 | **7.4279** | 10 |
| Style2[c] | **46.0319** | **0.9950** | **0.9978** | 7.3700 | 30 |

[a] Indicates that the content of the current image is a
[b] Indicates that the content of the current image is b

## IV. MATCHING RESULTS AND EVALUATION

### A. Evaluation index and test environment

In matching process, we consider the lack of richness of content and scarcity of targets in sonar images, the matching accuracy is used as the core of evaluation, *Ratio* is not set, cross-Check and RANSAC are turn on. The TFeat is combined with BRISK and SIFT detector separately for comparison. Fig. 4 depicts the difference in matching effects of a sonar image with subtle topography information before and after optical style transfer.

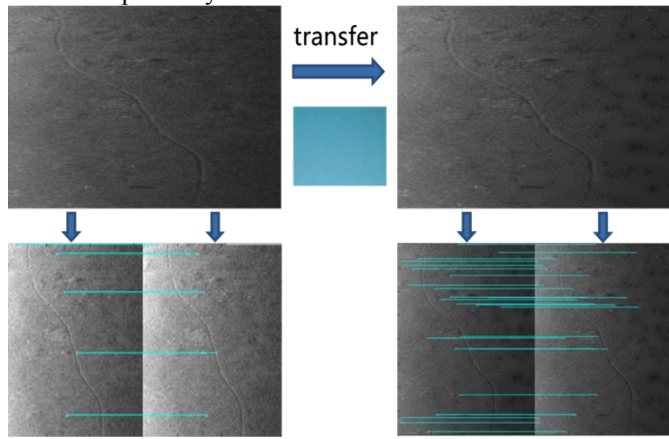

Figure 4. SIFT matching results before and after style transfer.

We introduce the max number of inliers (n_INLs), proportion of correct matches (POCM), which equals to the n_INLs divided by the sum of matches and running time (RT) as evaluation indexes, results are the average of ten tests. All methods were run under the Windows 10 operating system with an Intel Core i7-9700 3.00 GHz processor, 16 GB of physical memory and one NVIDIA GeForce RTX2070s graphics card. Sets of image pair 1-3 matching evaluation results are shown in the Table III. Tran indicates that the style transfer is complete.

TABLE III.  MATCHING EVALUATION RESULT 1 OF SONAR IMAGES

| Methods | n_INLs | | |
|---|---|---|---|
| | *Image pair1* | *Image pair2* | *Image pair3* |
| Raw+SIFT | 109 | 825 | 11 |
| Raw+BRISK | 120 | 154 | 8 |
| Tran+SIFT | 177 | 828 | **14** |
| Tran+BRISK | 161 | 171 | 8 |
| Ours[1] | 25 | **830** | 7 |
| Ours[2] | **309** | 176 | 7 |

[1] Indicates that combined with the SIFT detector
[2] Indicates that combined with the BRISK detector

TABLE IV.  MATCHING EVALUATION RESULT 2 OF SONAR IMAGES

| Methods | POCM | | |
|---|---|---|---|
| | *Image pair1* | *Image pair2* | *Image pair3* |
| Raw+SIFT | 0.171 | 0.683 | 0.687 |
| Raw+BRISK | 0.080 | 0.687 | 0.571 |
| Tran+SIFT | 0.191 | 0.683 | 0.875 |
| Tran+BRISK | 0.158 | 0.703 | 0.636 |
| Ours[1] | 0.072 | 0.760 | 0.778 |
| Ours[2] | **0.310** | **0.793** | **1.000** |

[1] Indicates that combined with the SIFT detector
[2] Indicates that combined with the BRISK detector

TABLE V.  MATCHING EVALUATION RESULT 3 OF SONAR IMAGES

| Methods | RT | | |
|---|---|---|---|
| | *Image pair1* | *Image pair2* | *Image pair3* |
| Raw+SIFT | 2.016 | 3.156 | 1.098 |
| Raw+BRISK | 0.347 | **0.243** | **0.354** |
| Tran+SIFT | 2.137 | 3.278 | 1.108 |
| Tran+BRISK | **0.343** | 0.251 | 0.358 |
| Ours[1] | 1.344 | 1.589 | 1.302 |
| Ours[2] | 2.101 | 1.405 | 1.409 |

[1] Indicates that combined with the SIFT detector
[2] Indicates that combined with the BRISK detector

Fig. 5 depicts different results of sonar images matching after transfer, including the SIFT, BRISK and ours. It is evident that for various underwater scenes, our approach achieves detecting more matching pairs while with minimal number of errors compared with the SIFT and BRISK.

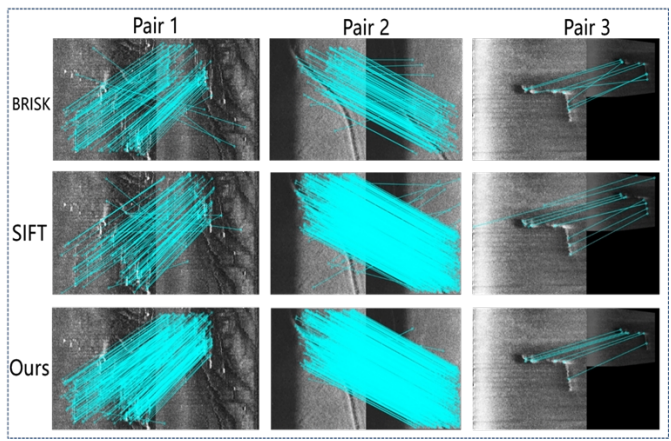

Figure 5. All the results of sonar images matching after style transfer.

## V. CONCLUSION

This paper uses the style transfer algorithm based on deep learning to preprocess the sonar images, then applies the learned descriptor TFeat to complete the matching task of underwater sonar images. It proves that the image after style transfer has content of sonar image and style of optical image. The style can significantly enhance the matching results. It should be emphasized that the TFeat can effectively solve the problem of matching sonar image pairs, although it has never been learned on sonar image datasets. Applying our proposed method to the underwater real scenes, it performs well in terms of accuracy, robustness and real-time performances.

In the future, we will further optimize the style transfer algorithm, reduce the complexity of custom parameters, and train the TFeat descriptor on underwater sonar images and equip it with a stronger detector to achieve better performances. Our research can provide a new idea for the task of sonar image matching. In addition, in the field of sonar image processing, the traditional methods often require sufficient professional knowledge, and the processing methods usually need complex overlay combinations. The style transfer method can be used as a new method of sonar image preprocessing.

ACKNOWLEDGMENT

The research was supported by the Chinese Shandong Provincial Key Research and Development Plan, under Grant No. 2019GHZ011. At the same time, thanks to the sonar image data support provided by PENG CHENG LABORATORY and Deep Vision AB company.